\documentclass[sigconf]{acmart}
\usepackage[utf8]{inputenc}
\AtBeginDocument{%
  }

\copyrightyear{2025}
\acmYear{2025}
\setcopyright{acmlicensed}
\acmConference[MRAC '25] {Proceedings of the 3rd International Workshop on Multimodal and Responsible Affective Computing}{October 27--31, 2025}{Dublin, Ireland.}
\acmBooktitle{Proceedings of the 3rd International Workshop on Multimodal and Responsible Affective Computing (MRAC '25), October 27--31, 2025, Dublin, Ireland}

\usepackage{amsmath}
\usepackage{algorithm}
\usepackage{algpseudocode}
\usepackage{graphicx}
\usepackage{hyperref}
\usepackage{subfig}
\usepackage{multirow}
\begin{document}

\title[Emotion Annotation Using Large Multimodal Models]{Zero-shot Emotion Annotation in Facial Images Using Large Multimodal Models: Benchmarking and Prospects for Multi-Class, Multi-Frame Approaches}

\author{He Zhang}
\email{hpz5211@psu.edu}
\orcid{0000-0002-8169-1653}
\affiliation{%
  \institution{Pennsylvania State University}
  \city{University Park}
  \state{PA}
  \country{USA}
}
\author{Xinyi Fu}
\orcid{0000-0001-6927-0111}
\authornote{Corresponding author}
\email{fuxy@tsinghua.edu.cn}
\affiliation{%
  \institution{Tsinghua University}
  \city{Beijing}
  \country{China}
}

\renewcommand{\shortauthors}{Zhang \& Fu}

\begin{abstract}
This study investigates the feasibility and performance of using large multimodal models (LMMs) to automatically annotate human emotions in everyday scenarios. We conducted experiments on the DailyLife subset of the publicly available FERV39k dataset, employing the GPT-4o-mini model for rapid, zero-shot labeling of key frames extracted from video segments. Under a seven-class emotion taxonomy (``Angry,'' ``Disgust,'' ``Fear,'' ``Happy,'' ``Neutral,'' ``Sad,'' ``Surprise''), the LMM achieved an average precision of approximately 50\%. In contrast, when limited to ternary emotion classification (negative/neutral/positive), the average precision increased to approximately 64\%. Additionally, we explored a strategy that integrates multiple frames within 1-2 second video clips to enhance labeling performance and reduce costs. The results indicate that this approach can slightly improve annotation accuracy. Overall, our preliminary findings highlight the potential application of zero-shot LMMs in human facial emotion annotation tasks, offering new avenues for reducing labeling costs and broadening the applicability of LMMs in complex multimodal environments.
\end{abstract}



\maketitle

\section{Introduction}
In the context of rapid advancements in artificial intelligence, technologies such as computer vision and natural language processing are being applied to a myriad of tasks to promote human well-being~\cite{obaigbena2024ai,le2021machine,calvo2014positive,cox1989national}. These technologies hold particular significance in providing emerging interaction methods within the field of human-computer interaction~\cite{wu2022survey}. They rely heavily on machine learning methods, where data annotation serves as a fundamental and indispensable step in model development ~\cite{zhou2017machine}.

In the development of machine learning models, data annotation serves as a foundational and indispensable step~\cite{wu2022survey, watson2024machine}. Accurate annotations are crucial for training machine learning models that can effectively interpret complex data, particularly in tasks involving human emotions and behaviors. However, the annotation process is notoriously labor-intensive and costly~\cite{zhou2017machine}, requiring annotators to spend prolonged periods meticulously labeling data~\cite{zhang2008active}. This manual effort not only demands significant human resources but also introduces variability and potential biases inherent in human cognition~\cite{
de2012contested,ding2022impact}. The challenge is magnified for emotion annotation tasks, where the subjective and nuanced nature of emotions complicates the labeling process. Addressing these challenges requires annotators to repeatedly review the data and engage in multiple rounds of iteration and discussion~\cite{10494076,7160695,schuff2017annotation}. 

To address these challenges, various annotation methodologies have been proposed, including the utilization of crowdsourcing platforms. Crowdsourcing can accelerate the annotation process by distributing the workload across a large number of annotators, thereby reducing both time and cost~\cite{vondrick2010efficiently}. Despite these advantages, crowdsourcing methods often prove insufficient when dealing with specialized environments or tasks that require nuanced understanding and expert judgment~\cite{10132377}. In such contexts, the reliance on human labor and expertise remains indispensable, highlighting the persistent need for more efficient and scalable annotation solutions.

Recent advancements in artificial intelligence (AI), particularly in the realm of large multimodal models (LMMs), have opened new avenues for automating annotation tasks~\cite{zhang2025augmenting,tan2024largelanguagemodelsdata}. LMMs, such as Generative Pre-trained Transformer (GPT), possess sophisticated natural language understanding capabilities and operate effectively in zero-shot settings, where they can perform tasks without explicit prior training on specific datasets~\cite{tan-etal-2024-large}. These models have demonstrated potential in various applications, from text generation to semantic understanding, suggesting their utility in assisting or even replacing human annotators~\cite{ZHANG2025100144,thapa2023humans,alizadeh2023open}.

Furthermore, the latest iterations of LMMs integrate visual capabilities, enabling them to comprehend and interpret graphical information in conjunction with textual data~\cite{10458347,shao2024automated}. This multimodal proficiency suggests that LMMs could become valuable tools for tasks that encompass both visual and linguistic components~\cite{10350935,10377575,10670574}. By leveraging their ability to understand visual inputs and operate in zero-shot settings, LMMs have the potential to streamline the annotation process while maintaining both accuracy and efficiency~\cite{10778143}. However, aside from capacity and performance considerations, the cost of computing resources is also very important. This matter is especially critical for individual users, students, small teams, and start-ups. Currently, because of the understandable business considerations, these groups rarely or even impossible to participate in the pricing process for computing resources such as API access and hardwares like graphic cards. Although we appreciate the discounts and grants offered to these groups, which show support from businesses and policies, these discounts do not reduce the overall use of computing resources. This means that their workflows and project goals remain limited by budget issues. In particular, some tasks or applications that depend heavily on computing resources may still not be easy to process.

Building on these capabilities and considering ways to reduce input and output costs, our study investigates the feasibility and performance of using LMMs for the automatic annotation of human emotions in everyday scenarios. Specifically, we employ the GPT-4o-mini model to conduct rapid, zero-shot labeling of key frames extracted from video segments within the DailyLife subset of the publicly available FERV39k dataset~\cite{Wang_2022_CVPR}. Our experiments assess the model's performance across two emotion taxonomies: a seven-class taxonomy encompassing ``Angry,'' ``Disgust,'' ``Fear,'' ``Happy,'' ``Neutral,'' ``Sad,'' and ``Surprise,'' and a ternary taxonomy categorizing emotions as negative, neutral, or positive.

Our results indicate that the LMM attained an average precision of approximately 50\% in the seven-class taxonomy, surpassing a simple baseline. This underscores the model's ability to discern complex emotional states without task-specific training. Notably, when the classification was constrained to a simpler ternary classification, the average precision increased to around 64\%, demonstrating the model's enhanced performance in broader emotion categories. Additionally, we investigated strategies of integrating multiple frames within 1-2 second video clips to improve labeling performance and reduce annotation costs. This approach resulted a slight but notable improvement in accuracy.

Our findings contribute to the growing body of research aimed at enhancing data annotation methods through AI. By examining the capabilities and limitations of zero-shot LMMs in emotion annotation tasks, we also discuss the potential for employing LMMs in practical applications to reduce costs and enhance scalability in real-world applications.

\vspace{-0.4cm}
\section{Related Work}
\subsection{Emotion Annotation}

Annotating human emotions has consistently been a challenging task~\cite{8764449}, not only due to the inherent complexity of emotions~\cite{5771357} but also because annotators may have varying evaluation standards (or subjectivity)~\cite{10416364,10150364}. A significant issue in emotion annotation is the annotation method. Although the most reliable annotation standard requires individuals to perform the annotations themselves, real-time self-annotation can lead to distraction and affect the expression of emotions~\cite{devillers2005challenges}. On the other hand, retrospective annotation relies on individuals' recollections~\cite{10.1145/2971485.2971516,7160695}, which may lead to bias~\cite{hoelzemann2024matter} as well as high cost~\cite{10.1145/3491102.3517453}, and can cause embarrassment~\cite{afzal2011natural}. Another widely used annotation method involves external annotators observing and labeling human emotions~\cite{8854185}. By leveraging human cognition and understanding, and considering the context, external annotators provide reliable emotional labels based on various cues~\cite{troiano2023dimensional}. Although these two annotation methods can be combined~\cite{10494076}, they may not be suitable for large-scale data processing. Regardless, emotion annotation remains a labor-intensive task. Therefore, exploring more efficient emotion annotation methods is crucial.

Currently, many annotation methods involve semi-automated or automated labeling conducted by models~\cite{10253654,10701433,10.1145/2912147,7112127,sharma2020automated,devillers2005challenges,9158345}, which greatly improves annotation efficiency. However, such annotations are typically built upon prior preparations, meaning that before the annotation task begins, data with emotion annotations are still required to train the underlying annotation models~\cite{you2016building}. Furthermore, in some specific tasks, these labels cannot be easily shared due to task constraints, but instead require the preparation of pre-trained data that suits specific scenarios~\cite{nimmi2022pre}. This implies that the traditional challenges in annotation tasks still persist.

Another important issue in emotion annotation is the choice of emotion classification scheme. Considering that annotation is a time-consuming and laborious process, researchers often categorize emotions based on task requirements to reduce the difficulty of annotation and improve efficiency. Examples include categorizing emotions into positive and negative~\cite{1034632}, emotional and neutral states~\cite{batliner2003find}, classifying specific emotions by their intensities~\cite{10494076}, and using basic emotions~\cite{10.1145/3629606.3629646,5585726}. In this study, we consider the potential task requirements of these various classification standards and base our research based on the available ground truth emotion labels.

\vspace{-0.2cm}
\subsection{LMM for Annotation}

The emergence and application of large language models (LLMs) and LMMs have introduced unprecedented opportunities in the field of data annotation. An increasing number of researchers and practitioners have recognized the vast potential of LLMs and LMMs for enhancing annotation processes~\cite{pmlr-v239-mohta23a}. As researchers continue to explore and leverage the advancing capabilities of LLMs, particularly in multimodal interactions~\cite{zhang2024mmllmsrecentadvancesmultimodal} and improvements in processing power~\cite{10.1145/3442188.3445922}, the range of annotation tasks has expanded significantly. These tasks now encompass various data types, including text \cite{10.1145/3637528.3671552}, audio \cite{10447760}, images \cite{cheng2024emotion, sapkota2024zero}, and specialized domain-specific data \cite{tang2024pdfchatannotator, zhang2024qualitativeresearchmeetslarge}.

A recent survey has shed light on current trends and leading research in the application of LMMs for annotation tasks \cite{tan-etal-2024-large}. Within the scope of our study, which focuses on emotion annotation for image data, related work has explored various capabilities of LMMs. For instance, researchers have evaluated the ability of LMMs to predict emotions from captions generated from images-derived captions~\cite{10388198}, perform image retrieval \cite{10.1145/3626772.3657740}, and generate descriptive captions \cite{shvetsova2025howtocaption}. Notably, in early 2024, a study compared the performance of LMMs such as GPT-3.5, GPT-4, and Bard against traditional supervised models like Convolutional Neural Networks (CNNs) for emotion recognition in image data \cite{nadeem2024vision}. The findings revealed that deep learning models specifically trained for this task generally achieved higher accuracy than LMMs.

However, despite the superior accuracy of traditional supervised models, they also present significant limitations. Nonetheless, LMMs offer the potential to achieve performance that is comparable to traditional models while reducing training and application costs. Therefore, in this study, we further optimized prompt engineering and reorganized annotation strategies to harness the capabilities and advantages of LMMs. 

\section{Method}

\subsection{Dataset}
We utilized the publicly available FERV39k dataset~\cite{Wang_2022_CVPR}, which comprises numerous 1-2 second video clips encompassing seven distinct emotions expressed by individuals across various scenarios (``Angry,'' ``Disgust,'' ``Fear,'' ``Happy,'' ``Neutral,'' ``Sad,'' ``Surprise''). This dataset has been manually annotated and extensively used in research, providing a widely recognized benchmark for comparative analyses. Within this dataset, we selected the ``DailyLife'' subset, as this scenario is considered the most representative of real-life conditions, thereby enhancing the potential transferability of our work to a broader range of task scenarios. Specifically, the ``DailyLife'' subset includes 2,339 video clips depicting a variety of daily activities, interactions, and emotional expressions. Each clip is manually annotated with a definitive emotion label based on contextual and visible emotional cues, serving as the ground truth label. Images were then extracted at 25 frames per second, each carrying the associated emotion label.

\subsection{Model Selection}
In this study, we employed the GPT-4o-mini (``gpt-4o-mini-2024-07-18'') model, a variant of the GPT-4 architecture optimized for greater efficiency and rapid inference. The selection of GPT-4o-mini was driven by its ability to perform zero-shot tasks while balancing performance~\footnote{\url{https://openai.com/index/gpt-4o-mini-advancing-cost-efficient-intelligence/}} and cost~\footnote{\url{https://openai.com/api/pricing/}} considerations. Additionally, GPT-4o-mini integrates vision capabilities~\footnote{\url{https://platform.openai.com/docs/guides/vision}}, allowing it to accept image inputs and interpret graphical information.

\subsection{Annotation Process}

\begin{figure*}[h!]
\centering
\includegraphics[width=0.9\textwidth]{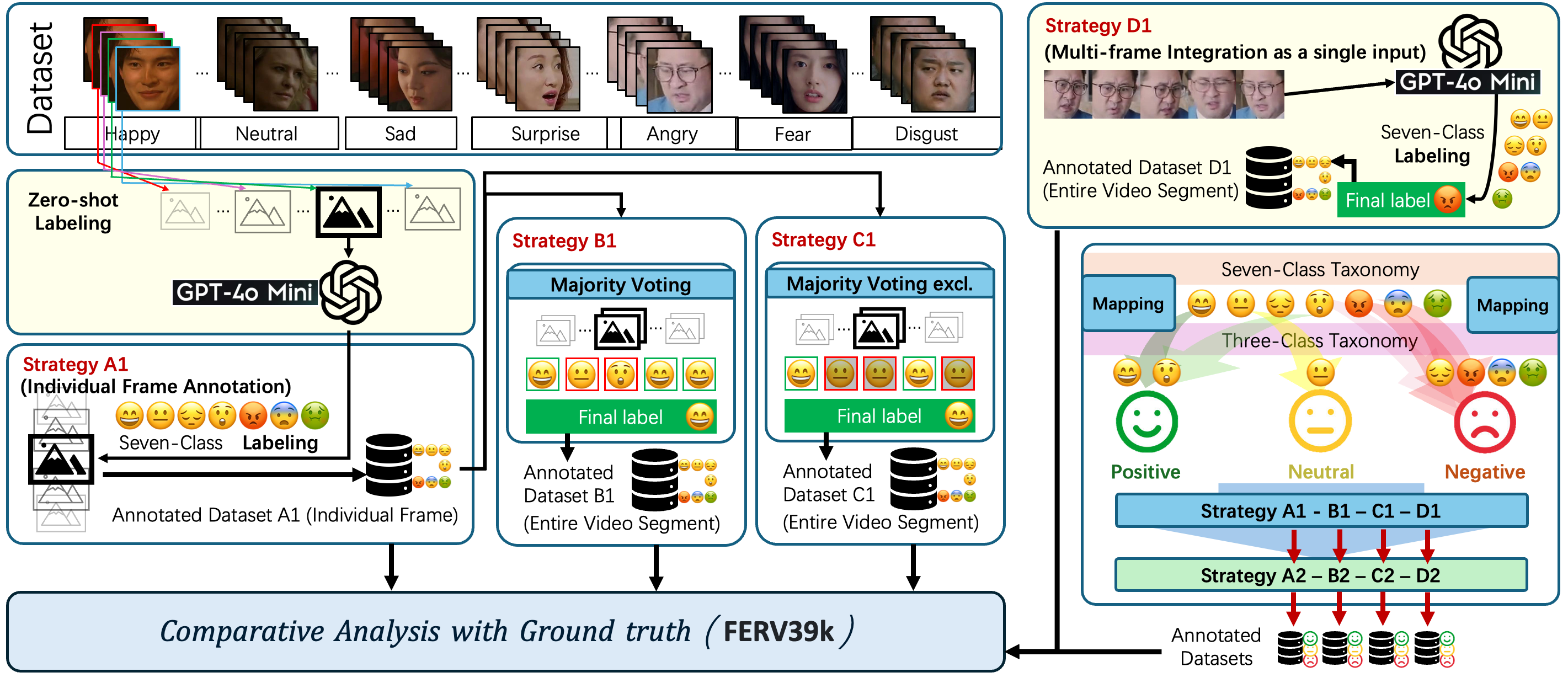}
\caption{Illustration of Multi-Strategy Annotation Framework for Emotion Recognition in Video (Image/Segment) Data}
\Description{The dataset consists of video frames categorized into seven emotion classes: happy, neutral, sad, surprise, angry, fear, and disgust. The process begins with zero-shot labeling using GPT-4o-mini, followed by four distinct annotation strategies, A1, B1, C1, and D1. Then, the annotations are mapped into both seven-class and three-class taxonomies (positive, neutral, negative), and proposed strategies A2, B2, C2, and D2. All results generated by strategies are processed for comparative analysis with ground truth dataset and models.}
\label{fig.framework}
\end{figure*}

The annotation methods and processes are illustrated in Fig~\ref{fig.framework}, with specific strategies to be detailed in the subsequent sections.

\subsubsection{Zero-Shot Labeling}

The annotation process used a zero-shot approach, prompting the GPT-4o-mini model with simple, predefined instructions to label extracted key frames. No additional training or fine-tuning was performed for the specific emotion annotation task. The prompts instructed the model to identify and classify the dominant emotion in each frame based on visual and contextual cues.

To optimize cost efficiency, five frames from each video segment were annotated: the initial frame, the first quartile (Q1), the middle frame, the third quartile (Q3), and the last frame. This sampling reduced annotation counts while capturing key emotional transitions. Different weighting strategies were then applied to derive comprehensive labels for each segment based on these frames. This approach balances accuracy in emotion recognition with cost constraints, ensuring the methodology remains effective and scalable. Apart from the predefined annotation strategy, we did not impose any additional human intervention on the LMM’s labeling process.

\subsubsection{Prompt Engineering}
In our study, we implemented prompt engineering to effectively utilize the GPT-4o-mini model for emotion annotation in images. The prompt was meticulously crafted to guide the model's responses by defining clear roles and providing specific instructions~\cite{lin2024write}. Initially, a prompt was set to establish the model as a ``\textit{professional image emotion analysis assistant},'' explicitly listing the available emotion options derived from the predefined $EMOTION\_LABELS$. This foundational setup ensures that the model operates within the desired context and understands the classification framework. For each image (or multi-frame integrated image) to be analyzed, we constructed a user message that includes both textual instructions and the image itself, e.g., ``\textit{This is an independent image frame, please analyze the emotion. Please analyze the emotion of the following image and select the most matching one from the above options, returning only the emotion name.}''

Subsequently, the user message was structured to include both textual and visual inputs. The textual component began with a customized prompt instructing the model to analyze the emotion conveyed in the image and select the most appropriate emotion from the provided options. This was followed by embedding the image itself, encoded in base64 format, within the message. By integrating the image (linked to local address) in this manner, we facilitated a multimodal interaction, allowing the model to process and interpret visual data alongside textual instructions.

The model was further directed to return only the name of the identified emotion, ensuring concise and relevant output. After the model generated a response, the content was extracted and stripped of any extraneous whitespace to obtain a clean final emotion label. In addition, we set the temperature parameter to 0 to ensure deterministic and consistent responses from the model.

\subsection{Annotation Strategies}

\subsubsection{\textbf{Annotation Strategy A1 (Seven-Class Taxonomy)}}\label{A1}

Strategy A1 involves individually annotating each of the five selected frames within a video segment using the seven-class emotion taxonomy. Specifically, the frames chosen for annotation include the initial frame, the first quartile (Q1) position frame, the middle frame, the third quartile (Q3) position frame, and the final frame of the segment. Each frame is independently labeled with one of the seven emotion categories: ``Angry,'' ``Disgust,'' ``Fear,'' ``Happy,'' ``Neutral,'' ``Sad,'' and ``Surprise.''

After annotation, the accuracy is directly calculated by comparing each frame's predicted emotion label against the ground truth labels provided in the dataset. 

\subsubsection{\textbf{Annotation Strategy B1 (Seven-Class Taxonomy)}}\label{B1}
Strategy B1 builds upon Strategy A1 by aggregating the emotion labels from the five annotated frames to determine the predominant emotion for the entire video segment. After individually annotating all five frames, the strategy identifies the absolute majority emotion among the labeled frames. In cases where there is a tie in the distribution of different emotions, the emotion label of the middle frame is selected to represent the video segment's overall emotional state.

\subsubsection{\textbf{Annotation Strategy C1 (Seven-Class Taxonomy)}}\label{C1}
Strategy C1 is determining the predominant emotion by excluding the ``Neutral'' category. Specifically, if one emotion constitutes an absolute majority among the annotated frames after removing ``Neutral,'' that emotion is assigned to the video segment. However, if all five frames are labeled as ``Neutral,'' the segment is assigned the ``Neutral'' label. In cases where there is an equal distribution of different emotions, the emotion label of the middle frame is selected to represent the overall emotion of the video segment. This approach aims to enhance annotation accuracy by focusing on more distinctly positive or negative emotional states, thereby mitigating the ambiguous property of LMM in classifying the intermediate emotion of ``neutral''.
\vspace{-0.2cm}
\subsubsection{\textbf{Annotation Strategy D1 (Seven-Class Taxonomy)}}\label{D1}
Strategy D1 employs a multi-frame integration approach by concatenating the five selected frames into a single composite input. Specifically, the initial frame, Q1 position frame, middle frame, Q3 position frame, and the final frames are sequentially joined along the temporal dimension, rather than across the channel, to form a unified image input. This consolidated input is then presented to the GPT-4o-mini model for annotation in a single step.

By integrating multiple frames, this strategy leverages temporal context, allowing the model to consider the emotional progression within the video segment. This holistic view aims to improve annotation accuracy by providing a broader context for emotion classification, potentially capturing transitional emotional states that individual frame annotations might miss.
\vspace{-0.2cm}
\subsubsection{\textbf{Annotation Strategy A2 (Three-Class Taxonomy)}}\label{A2}

Strategy A2 adapts the results from Strategy A1 to the three-class emotion taxonomy. In this strategy, each of the five annotated frames from Strategy A1 is directly mapped to one of three broader categories~\cite{Lee_2023_CVPR}: ``Positive,'' ``Neutral,'' or ``Negative.'' Specifically, emotions categorized as ``Angry,'' ``Disgust,'' ``Fear,'' and ``Sad'' are classified as ``Negative,'' while ``Happy'' and ``Surprise'' are classified as ``Positive.'' The ``Neutral'' labels are still ``Neutral''.

Each frame's seven-class label is converted to its corresponding three-class label based on this mapping. The accuracy is then calculated by comparing these three-class labels against the ground truth labels, allowing for an evaluation of the model's performance in a simplified emotion classification scenario.
\vspace{-0.2cm}
\subsubsection{\textbf{Annotation Strategy B2 (Three-Class Taxonomy)}}\label{B2}
Strategy B2 first applies Strategy A2 to reorganize seven-class labels into three classes. It then employs a strategy similar to Strategy B1, which returns the sentiment label with an absolute majority or uses the sentiment label of the middle frame if the sentiment trend scores are tied.
\vspace{-0.2cm}
\subsubsection{\textbf{Annotation Strategy C2 (Three-Class Taxonomy)}}\label{C2}
Strategy C2 involves first applying Strategy A2 to reorganize seven-class labels into three classes, followed by a strategy similar to Strategy C1 to mitigate the ambiguous property of LMM in classifying the intermediate emotion of ``neutral''.
\vspace{-0.2cm}
\subsubsection{\textbf{Annotation Strategy D2 (Three-Class Taxonomy)}}\label{D2}
Strategy D2 is similar to the multi-frame ensemble approach of Strategy D1, but uses a three-class classification approach. In this strategy, the five selected frames are concatenated into a single composite input, similar to Strategy D1. This integrated input is then processed by the GPT-4o-mini model to assign a single three-class emotion label (``Positive,'' ``Neutral,'' or ``Negative'') to the entire video segment.

\section{Results}

\subsection{Evaluation Metrics}

\begin{figure}[!htbp]
    \centering
    \subfloat[Precision Comparison Across Strategies for Seven-Class Annotation: Individual Metrics and Averages. This graph illustrates the precision scores for each emotion category across all strategies, along with macro and weighted averages denoted by dashed and dash-dot lines, respectively.]{
        \includegraphics[width=0.46\textwidth]{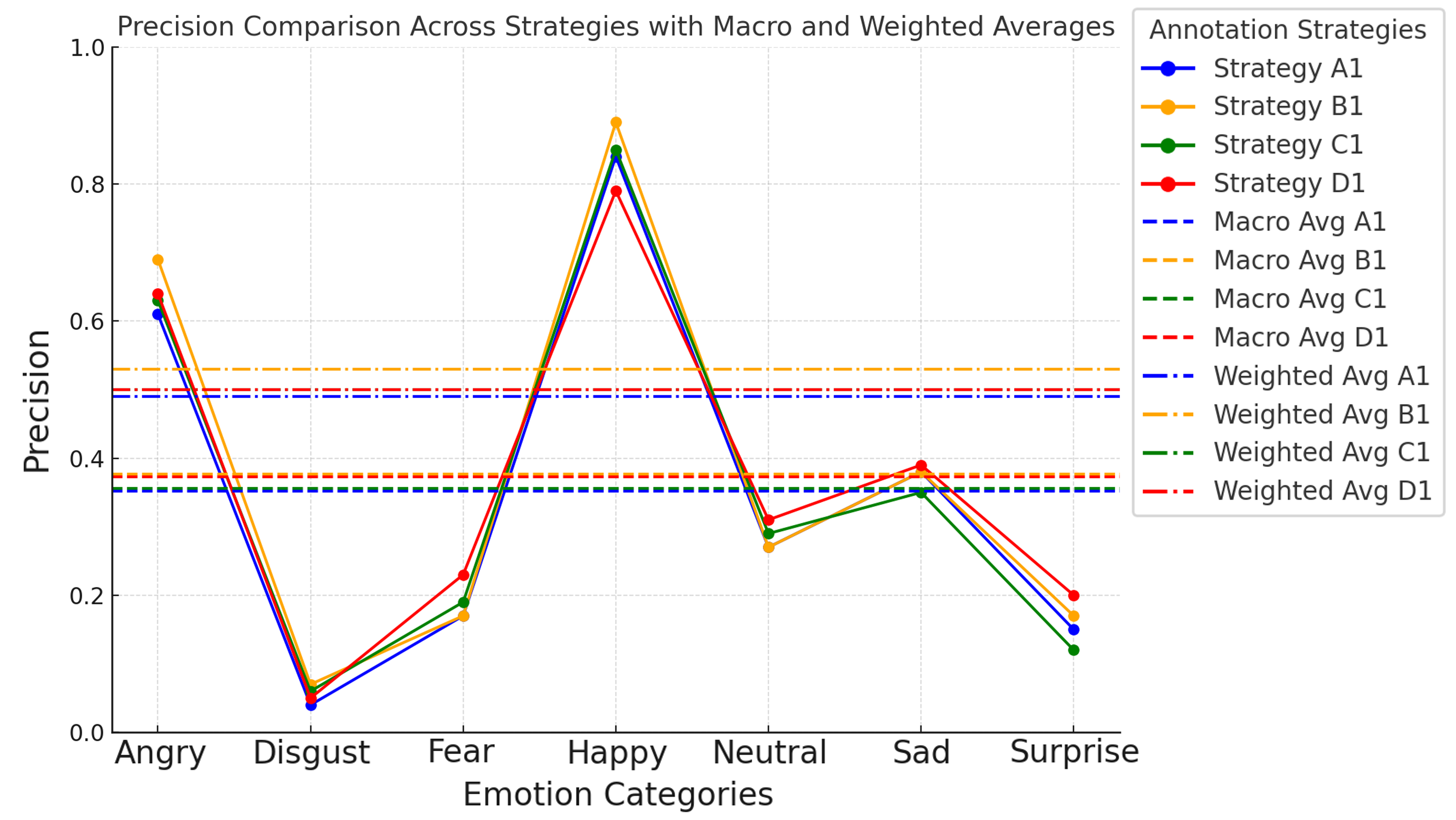}
        \label{fig:seven_class_graph}
    }
    \hfill
    \subfloat[Precision Comparison Across Strategies for Three-Class Annotation: Individual Metrics and Averages. This graph highlights the precision scores for each emotion category (negative, neutral, positive) across all strategies, alongside macro and weighted averages.]{
        \includegraphics[width=0.46\textwidth]{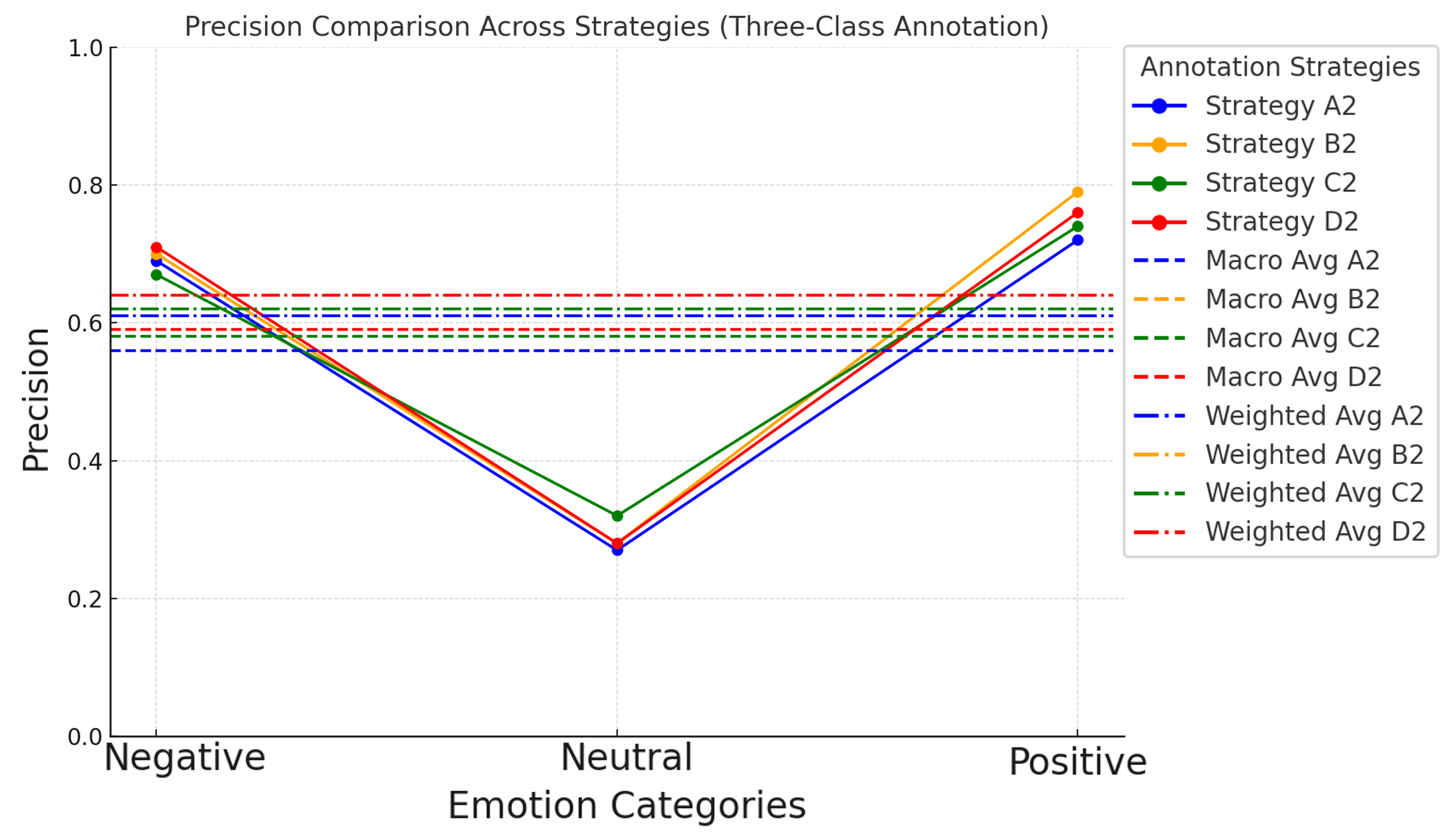}
        \label{fig:three_class_graph}
    }
    \caption{Precision Comparison for Seven-Class and Three-Class Annotation Strategies. Both graphs showcase individual metrics and overall averages (macro and weighted) for each strategy.}
    \label{fig:combined_graphs}
\end{figure}

We assessed our annotation strategies using precision, recall, F1-score, support, and accuracy (in Fig.~\ref{fig:combined_graphs}). \textbf{Precision} measures the proportion of correct predictions for each emotion, while \textbf{recall} evaluates the ability to identify all relevant instances. The \textbf{F1-score} balances precision and recall, making it useful for uneven class distributions. \textbf{Accuracy} reflects the overall correctness of the model. Additionally, we report \textbf{macro average} and \textbf{weighted average} to provide insights into performance across all classes, with macro average treating each class equally and weighted average accounting for class imbalance by weighting metrics based on class support. \textbf{Support} refers to the number of true instances for each emotion category in the dataset, providing context for the other metrics by indicating the distribution of classes.
\vspace{-0.8cm}
\setlength{\abovecaptionskip}{0.cm}
\vspace{-0.2cm}
\subsection{Seven-Class Taxonomy}
\begin{table*}[!]
\centering
\resizebox{\textwidth}{!}{%
\begin{tabular}{c|c|ccc|c|ccc|ccc|ccc}
\toprule\hline &
   &
  \multicolumn{3}{c|}{Annotation Strategy A1$^{\text{\ref{A1}}}$} &
   &
  \multicolumn{3}{c|}{Annotation Strategy B1$^{\text{\ref{B1}}}$} &
  \multicolumn{3}{c|}{Annotation Strategy C1$^{\text{\ref{C1}}}$} &
  \multicolumn{3}{c}{Annotation Strategy D1$^{\text{\ref{D1}}}$} \\ \hline
Label/Feature &
  support &
  precision &
  recall &
  f1-score &
  support &
  precision &
  recall &
  f1-score &
  precision &
  recall &
  f1-score &
  precision &
  recall &
  f1-score \\ \hline
Angry     & 3200  & 0.61 & 0.25 & 0.35 & 640  & 0.69 & 0.24 & 0.36 & 0.63 & 0.30 & 0.40 & 0.64 & 0.37 & 0.47          \\
Disgust   & 155   & 0.04 & 0.18 & 0.07 & 31   & 0.07 & 0.23 & 0.10 & 0.06 & 0.26 & 0.09 & 0.05 & 0.10 & 0.06          \\
Fear      & 690   & 0.17 & 0.12 & 0.14 & 138  & 0.17 & 0.12 & 0.14 & 0.19 & 0.14 & 0.16 & 0.23 & 0.22 & 0.22          \\
Happy     & 2415  & 0.84 & 0.39 & 0.53 & 483  & 0.89 & 0.42 & 0.57 & 0.85 & 0.55 & 0.66 & 0.79 & 0.62 & 0.70          \\
Neutral   & 2305  & 0.27 & 0.41 & 0.32 & 461  & 0.27 & 0.43 & 0.34 & 0.29 & 0.18 & 0.23 & 0.31 & 0.28 & 0.29          \\
Sad       & 2505  & 0.38 & 0.65 & 0.48 & 501  & 0.38 & 0.68 & 0.49 & 0.35 & 0.76 & 0.48 & 0.39 & 0.71 & 0.50          \\
Surprise  & 425   & 0.15 & 0.14 & 0.14 & 85   & 0.17 & 0.15 & 0.16 & 0.12 & 0.16 & 0.14 & 0.20 & 0.13 & 0.16          \\
accuracy  & 11695 &      &      & 0.38 &      &      &      & 0.40 &      &      & 0.41 &      &      & \textbf{0.46} \\
macro avg & 11695 & 0.35 & 0.30 & 0.29 & 2339 & 0.38 & 0.32 & 0.31 & 0.36 & 0.34 & 0.31 & 0.37 & 0.35 & 0.34          \\
weighted avg &
  11695 &
  0.49 &
  0.38 &
  0.39 &
  2339 &
  0.53 &
  0.40 &
  0.40 &
  0.50 &
  0.41 &
  0.41 &
  \textbf{0.50} &
  \textbf{0.46} &
  \textbf{0.46} \\ \hline\bottomrule
\end{tabular}%
}
\caption{Seven-Class annotation results using different annotation strategies. The relevant line graph is shown in Fig.~\ref{fig:seven_class_graph}.}
\label{tab:7class}
\end{table*}

Table~\ref{tab:7class} presents the performance metrics for four annotation strategies (A1, B1, C1, and D1) under the seven-class taxonomy. 

\textbf{Strategy A1 (Individual Frame Annotation) }attained an overall accuracy of 38\%. The model exhibited robust precision for the ``Happy'' category (0.84) but encountered significant challenges in accurately classifying ``Disgust'' (precision: 0.04). The recall metric was notably high for ``Sad'' (0.65) and considerably low for ``Disgust'' (0.18), highlighting the model's difficulty in reliably identifying certain emotional states.

\textbf{Strategy B1 (Majority Voting)} yielded an incremental improvement in accuracy, reaching 41\%. Precision for ``Happy\\ rose to 0.89, while ``Disgust''
experienced marginal enhancements in both precision (0.07) and recall (0.23). This suggests that aggregating frame-level annotations through majority voting can slightly bolster performance for specific emotions.

\textbf{Strategy C1 (Majority Voting Excluding ``Neutral'')} further augmented the accuracy to 46\%. By excluding the ``Neutral'' category from the majority voting process, this approach improved recall for ``Sad'' to 0.76 and maintained high precision for ``Happy'' (0.85). This indicates that focusing on ``Negative'' and ``Positive'' emotions can mitigate some inaccuracies associated with the ``Neutral'' classifications, thereby enhancing overall annotation reliability.

\textbf{Strategy D1 (Multi-Frame Integration)} achieved an accuracy of 46\%, paralleling Strategy C1. By amalgamating multiple frames into a single input, this strategy effectively harnessed temporal context, thereby improving the model's capacity to capture the dynamic progression of emotions across video segments. This integration allows the model to consider the emotional transitions and consistencies present within the selected frames, leading to more coherent and accurate segment-level annotations.

Additionally, when considering the macro average and weighted average metrics, Strategies C1 and D1 not only achieved higher accuracy but also demonstrated improved balanced performance across all classes. The \textbf{macro average} indicates that these strategies perform more consistently across less frequent emotion categories, while the \textbf{weighted average} reflects their enhanced overall performance, accounting for class imbalances in the dataset.

\begin{table}[ht]
\centering
\begin{tabular}{lcccc}
\toprule\hline
\textbf{Source} & \textbf{W} & \textbf{df} & \textbf{Q} & \textbf{$p$-unc} \\
\midrule
Method          & 0.0129    & 2         & 60.37    & $7.78\times10^{-14}$ \\
\hline\bottomrule
\end{tabular}
\caption{Friedman Test Results for Merged Data (Strategy B1, C1, and D1). This table summarizes the outcome of a Friedman test applied to compare the prediction correctness across the merged dataset for three methods. The test indicates that there is a statistically significant difference in performance among the methods.}
\label{tab:friedman}
\end{table}


\begin{table}[ht]
\centering
\begin{tabular}{lccc}
\toprule
\hline
                  & \textbf{Correct\_B1} & \textbf{Correct\_C1} & \textbf{Correct\_D1} \\
\midrule
\textbf{Correct\_B1} & 1.000000           & 1.000000           & 0.000218           \\
\textbf{Correct\_C1} & 1.000000           & 1.000000           & 0.005641           \\
\textbf{Correct\_D1} & 0.000218           & 0.005641           & 1.000000           \\
\hline\bottomrule
\end{tabular}
\caption{Dunn Post-Hoc Test Results (Bonferroni Adjusted) for Merged Data. Pairwise comparisons using Dunn's test reveal that while two of the Strategies (B1 and C1) do not differ significantly, the Strategy D1 shows a statistically significant difference from Strategy B1 and C1.}
\label{tab:dunn}
\end{table}

To further validate these observations, a non-parametric Friedman test was conducted on the merged dataset (comprising the results from Strategies B1, C1, and D1). As shown in Table~\ref{tab:friedman}, the Friedman test revealed a statistically significant difference among these strategies (p<0.05). This overall significance suggests that at least one strategy performs differently compared to the others.

Subsequently, pairwise comparisons were performed using Dunn’s post-hoc test with Bonferroni correction (see Table~\ref{tab:dunn}). The results indicate that there is no statistically significant difference between Strategies B1 and C1. In contrast, Strategy D1 is statistically significantly different from both B1 and C1 (p<0.05). These statistical findings provide robust evidence that the multi-frame integration approach (Strategy D1), which harnesses temporal context, exhibits a distinct performance profile compared to the majority voting methods.

Furthermore, when considering the macro average and weighted average metrics, both Strategies C1 and D1 demonstrated improved balanced performance across all emotion categories. The macro average reflects consistency in performance across less frequent categories, while the weighted average accounts for class imbalances, further reinforcing the overall effectiveness of these strategies.
\vspace{-0.2cm}
\subsection{Three-Class Taxonomy}

\begin{table*}[!]
\centering
\resizebox{\linewidth}{!}{%
\begin{tabular}{c|c|ccc|c|ccc|ccc|ccc}
\toprule\hline
 &
   &
  \multicolumn{3}{c|}{Annotation Strategy A2$^{\text{\ref{A2}}}$} &
   &
  \multicolumn{3}{c|}{Annotation Strategy B2$^{\text{\ref{B2}}}$} &
  \multicolumn{3}{c|}{Annotation Strategy C2$^{\text{\ref{C2}}}$} &
  \multicolumn{3}{c}{Annotation Strategy D2$^{\text{\ref{D2}}}$} \\ \hline
Label/Feature &
  support &
  precision &
  recall &
  f1-score &
  support &
  precision &
  recall &
  f1-score &
  precision &
  recall &
  f1-score &
  precision &
  recall &
  f1-score \\ \hline
negative     & 6549  & 0.69 & 0.70 & 0.69 & 1310 & 0.70 & 0.74      & 0.72      & 0.67 & 0.87      & 0.76 & 0.71          & 0.80      & 0.75          \\
neutral      & 2305  & 0.27 & 0.41 & 0.32 & 461  & 0.28 & 0.41      & 0.33      & 0.32 & 0.16      & 0.22 & 0.31          & 0.28      & 0.29          \\
positive     & 2840  & 0.72 & 0.39 & 0.50 & 568  & 0.79 & 0.41      & 0.54      & 0.74 & 0.53      & 0.62 & 0.76          & 0.58      & 0.66          \\
accuracy     & 11694 &      &      & 0.57 & 2339 &      & \textbf{} & \textbf{} &      & \textbf{} & 0.65 & \textbf{}     & \textbf{} & \textbf{0.65} \\
macro avg    & 11694 & 0.56 & 0.50 & 0.51 & 2339 & 0.59 & 0.52      & 0.53      & 0.58 & 0.52      & 0.53 & 0.59          & 0.55      & 0.57          \\
weighted avg & 11694 & 0.61 & 0.57 & 0.58 & 2339 & 0.64 & 0.59      & 0.60      & 0.62 & 0.65      & 0.62 & \textbf{0.64} & \textbf{0.65}      & \textbf{0.64} \\ \hline\bottomrule
\end{tabular}%
}
\caption{Three-Class annotation results using different annotation strategies. The relevant line graph is shown in Fig.~\ref{fig:three_class_graph}.}
\label{tab:3class}
\end{table*}

Table~\ref{tab:3class} presents the performance metrics for four annotation strategies within the three-class taxonomy framework.

\textbf{Strategy A2 (Mapped Three-Class Classification)} achieved an accuracy of 57\%. The ``Positive'' category exhibited strong precision (0.72), whereas the ``Neutral'' category demonstrated moderate performance with a precision of 0.27 and recall of 0.41.

\textbf{Strategy B2 (Majority Voting)} resulted in a substantial accuracy improvement, attaining 65\%. Precision for ``Positive'' increased to 0.79, while the ``Negative'' category demonstrated robust performance with a precision of 0.70 and recall of 0.74.

\textbf{Strategy C2 (Majority Voting Excluding ``Neutral'')} also achieved an accuracy of 65\%. This strategy maintained high precision for ``Negative'' (0.67) and significantly improved recall for ``Negative'' to 0.87, while the ``Positive'' category maintained consistent performance with a precision of 0.76 and recall of 0.58.

\textbf{Strategy D2 (Multi-Frame Integration)} matched the accuracy of 65\%, effectively leveraging both temporal context and simplified emotion categories to ensure efficient and accurate annotation.

\begin{table}[ht]
\centering
\caption{Friedman Test Results for Merged Data (Methods B2, C2, and D2).}
\label{tab:friedman9}
\begin{tabular}{lcccc}
\toprule\hline
\textbf{Source} & \textbf{W} & \textbf{df} & \textbf{Q} & \textbf{$p$-unc} \\
\midrule
Method & 0.014 & 2 & 67.178 & $2.585\times10^{-15}$ \\
\hline\bottomrule
\end{tabular}
\end{table}

\begin{table}[ht]
\centering
\caption{Dunn Post-Hoc Test Results (Bonferroni Adjusted) for Merged Data.}
\label{tab:dunn9}
\begin{tabular}{lccc}
\toprule\hline
                 & \textbf{Correct\_B2} & \textbf{Correct\_C2} & \textbf{Correct\_D2} \\
\midrule
\textbf{Correct\_B2} & 1.000000           & 0.000195           & 0.000251           \\
\textbf{Correct\_C2} & 0.000195           & 1.000000           & 1.000000           \\
\textbf{Correct\_D2} & 0.000251           & 1.000000           & 1.000000           \\
\hline\bottomrule
\end{tabular}
\end{table}

Overall, \textbf{Strategies B2}, \textbf{C2}, and \textbf{D2} consistently outperformed \textbf{Strategy A2}, further highlighting the effectiveness of aggregation and integration methods in enhancing annotation accuracy within zero-shot classification approaches based on LMMs. Furthermore, the \textbf{macro average} and \textbf{weighted average} metrics underscore the balanced performance of \textbf{Strategies B2, C2, and D2} across all emotion categories. The macro average indicates that these strategies maintain consistent precision and recall across both common and rare classes, while the weighted average reflects their strong overall performance by accounting for the distribution of classes in the dataset.

To further validate these observations, a non-parametric Friedman test was conducted on the merged dataset (comprising the results of Strategies B2, C2, and D2). As summarized in Table~\ref{tab:friedman9}, the Friedman test revealed a statistically significant difference among the three methods (p < 0.05), which supports the notion that at least one strategy produces a performance profile that is distinct from the others. Subsequent pairwise comparisons using Dunn’s post-hoc test with Bonferroni adjustment (see Table~\ref{tab:dunn9}) demonstrated that Strategy B2 significantly differs from both Strategies C2 and D2 (p < 0.05), while no significant difference was found between Strategies C2 and D2. This statistical evidence reinforces our conclusion that the choice of aggregation strategy plays a critical role in determining annotation accuracy and consistency.

In summary, the combined performance metrics and the statistical tests confirm that the aggregation and integration methods (especially Strategies C2 and D2) not only yield superior overall accuracy but also provide a more balanced performance across diverse emotion categories, thereby underscoring the importance of temporal context and strategic data integration in zero-shot annotation tasks.
\vspace{-0.2cm}
\subsection{Performance of Different Strategies: Insights from Confusion Matrices}

The presented confusion matrices compare the performance of various classification strategies (A1, A2, B1, B2, C1, C2, D1, D2) across different tasks involving emotion and sentiment recognition. Each matrix visualizes the true labels versus the predicted labels, providing insights into the model's accuracy, strengths, and areas requiring improvement. Strategies A1, B1, C1, and D1 are evaluated on their ability to classify seven distinct emotional states, including ``Angry,'' ``Happy,'' ``Neutral,'' and ``Sad.'' The diagonal entries reflect correct classifications, while off-diagonal values indicate confusion between emotions. For example, significant misclassifications are observed between ``Neutral'' and ``Happy'' in some strategies, highlighting challenges in distinguishing subtle emotional variations. Strategies A2, B2, C2, and D2 focus on sentiment classification into three categories: ``Negative,'' ``Neutral,'' and ``Positive.'' While these strategies generally achieve high accuracy for the ``Negative'' category, frequent confusion between ``Neutral'' and ``Positive'' suggests a need for improved sensitivity to nuanced sentiment expressions.

Different strategies exhibit similar distribution patterns, indicating that the zero-shot LMM annotation approach introduces a certain level of ambiguity in labeling the aforementioned emotion categories. This ambiguity often leads to confusion between certain categories, while labels for negative emotions are generally more accurate. This observation suggests potential opportunities for emotion recognition tasks in specific contexts, where leveraging the strengths of zero-shot LMM annotation could enhance performance despite its inherent limitations.

\subsection{Comparative Analysis}

Comparing the seven-class and three-class taxonomies, it is evident that simplifying emotion classification enhances overall accuracy. The three-class strategies (B2, C2, D2) achieved an accuracy of 65\%, significantly higher than the best seven-class strategies (C1 and D1) at 46\%. This improvement is attributed to the reduced complexity in classification, allowing the model to more effectively distinguish between "Negative," "Neutral," and "Positive" emotions.

Furthermore, aggregation methods — whether through majority voting (B1, B2, C1, C2) or multi-frame integration (D1, D2) — consistently yielded better performance compared to individual frame annotation (A1, A2). These findings highlight the importance of leveraging temporal context and strategic frame selection to enhance the reliability and accuracy of automated emotion annotation.

Overall, the results demonstrate that the GPT-4o-mini model is capable of effectively annotating human emotions, particularly when employing strategies that aggregate information from multiple frames and simplify emotion categories. These approaches offer a balanced trade-off between annotation accuracy and computational efficiency, making them suitable for large-scale, real-world applications.

\subsubsection{Baseline Comparison (Random Guessing)}

To contextualize the performance of our annotation strategies, we compared them against baseline theoretical chance levels derived from random guessing. In the seven-class taxonomy, random guessing would yield an expected accuracy of approximately 14.3\%, while in the three-class taxonomy, the expected accuracy is around 33.3\%. Our results demonstrate that all proposed strategies significantly surpass these baseline levels. Specifically, in the seven-class taxonomy, the best-performing strategies (C1 and D1) achieved an accuracy of 46\%, more than three times the baseline. Similarly, in the three-class taxonomy, Strategies B2, C2, and D2 reached an accuracy of 65\%, nearly doubling the random guessing baseline. This substantial improvement underscores the effectiveness of our aggregation and integration methods in enhancing annotation accuracy within zero-shot classification tasks using large language models.

\subsubsection{Baseline Comparison (Pre-trained Models and Zero-shot Methods)}

To further contextualize the performance of our annotation strategies, we compared our results against baseline models reported in the FERV39k dataset paper \cite{Wang_2022_CVPR}, with a particular focus on the DailyLife subset under the seven-class taxonomy. The baseline models encompass various architectures, including ResNet-18 (R18), ResNet-50 (R50), VGG-13 (VGG13), VGG-16 (VGG16), and their LSTM-enhanced variants. The performance metrics reported are WAR (Weighted Average Recall)\footnote{WAR measures the average recall across all classes, weighted by the number of true instances in each class, thereby emphasizing performance on more frequent classes.} and UAR (Unweighted/Macro Average Recall)\footnote{UAR calculates the average recall without weighting, treating each class equally regardless of its frequency, which ensures that the model's performance on minority classes is adequately represented.}, which provide a balanced evaluation by accounting for class imbalances and ensuring that each class contributes proportionally to the overall performance. 

In the DailyLife category, baseline models achieved the following WAR/UAR scores as shown in Table~\ref{tab:comparison}. 

\begin{table}[!htbp]
\centering
\begin{tabular}{llcc}
\hline
\textbf{Category} & \textbf{Method}                      & \textbf{WAR (\%)} & \textbf{UAR (\%)} \\ 
\hline
\multirow{17}{*}{\textbf{Baseline}~\cite{Wang_2022_CVPR}} 
& R18                                  & 41.40            & 31.13             \\
& R50                                  & 31.00            & 19.37             \\
& VGG13                                & 39.07            & 28.63             \\
& VGG16                                & 41.19            & 28.73             \\
& R18-LSTM                             & 41.61            & 29.11             \\
& R50-LSTM                             & 41.61            & 28.00             \\
& VGG13-LSTM                           & 46.07            & 31.50             \\
& VGG16-LSTM                           & 44.37            & 30.58             \\
& C3D~\cite{Tran_2015_ICCV}           & 26.96            & 18.35             \\
& I3D~\cite{Carreira_2017_CVPR}       & 39.70            & 26.09             \\
& 3D-R18~\cite{Tran_2018_CVPR}        & 35.67            & 24.95             \\
& Two C3D                              & 35.46            & 23.26             \\
& Two I3D                              & 40.76            & 28.93             \\
& Two 3D-R18                           & 39.28            & 28.41             \\
& Two R18-LSTM                         & 40.55            & 27.09             \\
& Two VGG13-LSTM                       & 46.92            & 31.55             \\
& AEN~\cite{Lee_2023_CVPR} & 47.88 & 38.18                 \\
& \textbf{Average*}                     & \textbf{39.97}   & \textbf{27.87}    \\ \cline{2-4} 
& CLIP~\cite{pmlr-v139-radford21a}$^{\beta}$ & 17.10  & 20.99    \\
& FaRL~\cite{Zheng_2022_CVPR}$^{\beta}$ & 25.65 & 21.67           \\
& EmoCLIP~\cite{10581982}$^{\beta}$ & 35.30 & 26.73                \\ 
& \textbf{Average}$^{\beta}$   & \textbf{26.02}   & \textbf{23.13}    \\ \cline{2-4} 
& FineCLIPER~\cite{10.1145/3664647.3680827}$^{\gamma}$ & 48.63 & 40.79                 \\ \cline{2-4} 
& \textbf{Average}                     & \textbf{38.39}   & \textbf{27.81}    \\ 
\hline
\textbf{This Study} & \textbf{Zero-Shot LMM}$^{\alpha}$ & \textbf{46.00}    & \textbf{35.00}    \\
\hline
\multicolumn{4}{l}{*Pre-trained}   \\
\multicolumn{4}{l}{$^{\alpha}$Built on GPT-4o-mini }   \\
\multicolumn{4}{l}{$^{\beta}$The result of zero-shot method}   \\
\multicolumn{4}{l}{$^{\gamma}$The result of zero-shot method with label augmentation}   \\
\end{tabular}
\caption{Comparison of Weighted Average Recall (WAR) and Unweighted Average Recall (UAR) between baseline architectures (trained from scratch) and the LMM-based zero-shot method (GPT-4o-mini) on the FERV39k DailyLife subset~\cite{Wang_2022_CVPR}.}
\label{tab:comparison}
\end{table}


Our best-performing strategy within the seven-class taxonomy, Strategy D1 (Multi-Frame Integration), achieved a WAR of 46\%, which closely rivals top-performing pre-trained baseline methods such as AEN (47.88\% WAR), VGG13-LSTM (46.07\% WAR) and Two VGG13-LSTM (46.92\% WAR), while significantly exceeding the baseline average of approximately 38.98\% WAR. Moreover, Strategy D1 significantly surpasses the average WAR of the baseline models, which stands at approximately 38.98\%. Compared to other zero-shot methods, our Strategy D1's result are higher than the results of zero-shot methods. Although FineCLIPER (48.63\% WAR and 40.79\% UAR) achieves better performance, it utilizes additional label augmentation, which we have not employed. We suggest that future research could explore a combination of FineCLIPER’s and our methods, aiming to further save computational resources while still improving recognition results.

\section{Cost-Efficiency and Scalability}
Besides performing close to or outperforming baseline models in performance, compared to traditional supervised models, GPT-4o-mini-based annotation strategies have significant advantages in cost-effectiveness and scalability. Strategy D1 reduces operational costs by minimizing the number of API requests and decreasing token lengths through preprocessing. This cost-effective approach ensures that large-scale annotation tasks remain financially feasible. Furthermore, our zero-shot annotation approach leverages the capabilities of LMMs without necessitating task-specific training, allowing for rapid deployment and adaptation to various annotation tasks with minimal additional resources.

While some baseline models, such as Two VGG13-LSTM, exhibit marginally higher WAR scores, our annotation strategy D1 achieves comparable performance levels with enhanced cost and operational efficiencies. This underscores the effectiveness of aggregation techniques and temporal context integration in zero-shot annotation tasks using LMMs, presenting a viable alternative to traditional supervised models, especially in scenarios constrained by budget and computational resources.

\vspace{-0.2cm}
\section{Conclusion with Future Work}
This study demonstrates the feasibility of using LMMs for automated emotion annotation in facial images through zero-shot classification. By exploring various annotation strategies, we identified the potential of LMMs to achieve competitive performance, particularly in tasks involving ternary classification of emotions. Strategies that integrate multiple frames or aggregate annotations through majority voting significantly enhance the reliability of emotion recognition, offering a promising alternative to traditional supervised methods.

While LMMs exhibit inherent ambiguity in distinguishing closely related emotion categories, particularly within the seven-class taxonomy, they achieve higher accuracy in simpler classification tasks. This highlights their utility in scenarios where efficiency and scalability are prioritized over fine-grained classification precision. Moreover, the cost-effective nature of zero-shot LMM annotation enables large-scale deployment, reducing the reliance on human annotators and minimizing operational costs.

Our findings underscore the importance of leveraging aggregation techniques, temporal context, and task simplification to maximize the potential of LMMs in emotion annotation. Future work should focus on fine-tuning multimodal LMMs for emotion recognition, addressing ambiguities in classification, and expanding their application in real-world multimodal environments, such as driver attention detection, live streaming platform moderation,  and health management systems. This research provides a foundation for advancing automated annotation techniques, fostering innovation in human-computer interaction and affective computing domains.

\section*{Safe and Responsible Innovation Statement}
We hope to take this opportunity to remind researchers employing this method to use it responsibly and thoughtfully, particularly under the concept of superalignment. It is also important to note that LLMs may hallucinate in the label space, potentially introducing inaccurate or spurious outputs that could mislead downstream processes. In this paper, our tests were carried out by calling the API, and under the privacy policy, none of these data will be shared or used for training purposes. In our study, the testing was carried out via API calls; under the privacy policy, this data will neither be shared nor used for training purposes. We also recommend experimenting with on-premise models in the future to eliminate the risk of data leaks. Beyond that, when prototyping responsibly around this capability, it may be wise to target specific tasks and scenarios first rather than rolling it out directly to a wider market. 


\begin{acks}
We thank the anonymous reviewers for their valuable feedback. We also acknowledge the authors of the FERV39k dataset for their contributions and for granting permission for its non-commercial use. This work was supported by the Beijing Natural Science Foundation-Youth Project (Grant No.4254082).
\end{acks}

\bibliographystyle{ACM-Reference-Format}
\balance
\bibliography{sample-base}

\end{document}